\documentclass[letterpaper]{article} 
\usepackage{aaai2027}  
\usepackage{amsmath}
\usepackage{amssymb}
\usepackage[hyphens]{url}  
\usepackage{graphicx} 
\usepackage{natbib}  
\usepackage{caption} 
\usepackage{algorithm}
\usepackage{algorithmic}

\usepackage{newfloat}
\usepackage{listings}
\DeclareCaptionStyle{ruled}{labelfont=normalfont,labelsep=colon,strut=off} 
\floatstyle{ruled}
\newfloat{listing}{tb}{lst}{}
\floatname{listing}{Listing}

\usepackage{booktabs}

\title{SpIn-ViT: Designing a Sparsity-Induced Vision Transformer That Is Mechanistically Interpretable}
\author{
    Philip H. Lee\textsuperscript{\rm 1}, Parth Padalkar\textsuperscript{\rm 2}
}
\affiliations{
    \textsuperscript{\rm 1} Independent Researcher\\
    \textsuperscript{\rm 2} Texas State University\\
    leephiliphl@gmail.com, padalkar@txstate.edu

}

\begin{document}

\maketitle

\begin{abstract}
Mechanistic interpretability has recently expanded to Vision Transformers (ViTs), with Sparse Autoencoders (SAEs) increasingly used as post-hoc tools to decompose internal representations into sparse and more interpretable features. However, because post-hoc SAEs are trained on frozen representations after the ViT has already been optimized, their latent features are not directly aligned with the downstream classification objective. We introduce SpIn-ViT, a framework that jointly trains a pretrained ViT and a modified SAE end-to-end, directly aligning sparse patch-level representations with image classification. SpIn-ViT learns semantically coherent neuron activations that localize meaningful image regions while maintaining competitive predictive performance. We evaluate SpIn-ViT across nine image-classification benchmarks using classification accuracy, quantitative interpretability metrics, AI-based and Human evaluations. Compared with the previous state-of-the-art post-hoc SAE method, SpIn-ViT achieves \textbf{8.84\%} higher average classification accuracy, an AI-based interpretability score nearly four times as high, and a human-evaluation score more than twice as high. We further extract interpretable rule-sets using the SAE neurons to create neurosymbolic models which achieve \textbf{5.97\%} higher average classification accuracy while requiring a \textbf{58.85\%} smaller rule-set than the neurosymbolic models created from the SOTA post-hoc SAE method.
\end{abstract}


\begin{figure*}[t]
    \centering
    \includegraphics[height=6cm,width=1.0\textwidth]{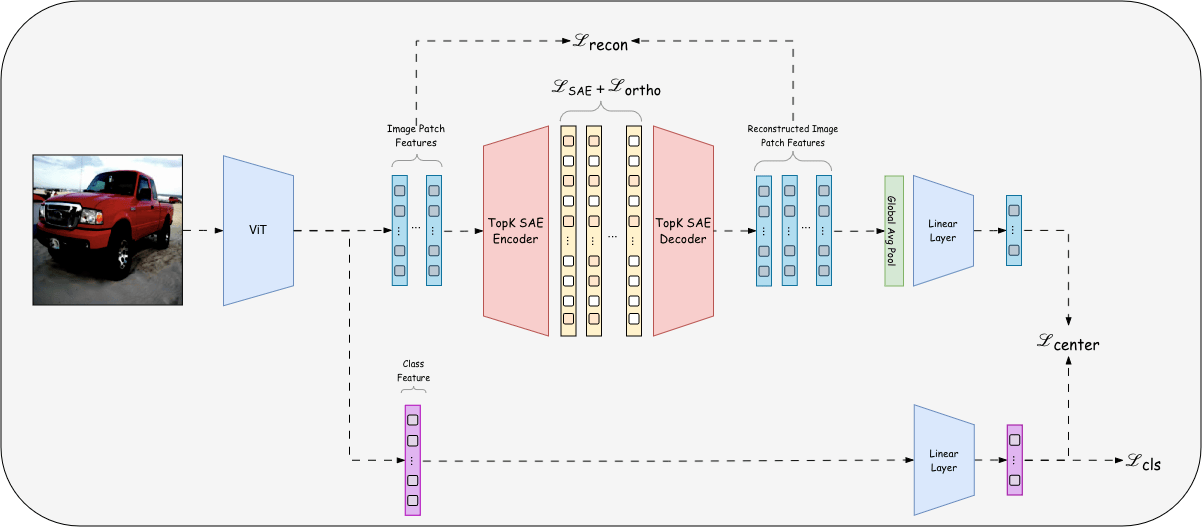}
    \caption{Training SpIn-ViT. ViT patch embeddings are encoded into a sparse latent space, reconstructed through an SAE decoder, and aligned with CLS-token predictions using reconstruction, sparsity, and alignment losses.}
    \label{fig:SpIn-ViT-Training}
\end{figure*}

\section{Introduction}

Transformers have been adapted to images, providing a powerful backbone for modern computer vision \cite{dosovitskiy2021animage}. By processing images as patch tokens through multi-head self-attention, Vision Transformers (ViTs) offer a general framework that performs well across diverse tasks, including image classification \cite{dosovitskiy2021animage, touvron2021training}, object detection \cite{zhu2020deformabledetr, carion2020endtoend}, and image segmentation \cite{strudel2021segmenter, xie2021segformer}. However, their internal mechanisms remain incompletely understood, motivating growing interest in methods for interpreting their predictions.

In ViTs, multiple concepts may be entangled within similar internal representations, giving rise to polysemantic features that are difficult to interpret. Furthermore, ViTs are commonly fine-tuned using only the class token. Consequently, patch-level representations receive no direct supervision from the classification objective, making it difficult to extract interpretable features from them in isolation. These challenges motivate the study of ViTs through the lens of mechanistic interpretability.

Mechanistic interpretability seeks to understand how machine learning models arrive at their predictions by analyzing their internal representations and computations \cite{DBLP:journals/corr/abs-2501-16496}. A prevalent approach is to use Sparse Autoencoders (SAEs) \cite{sae} to disentangle polysemantic features into more monosemantic components that are easier to interpret. Extensive work has applied SAEs as post-hoc interpretability methods for large language models (LLMs) \cite{chughtai2023sparse, bricken2023towards, rajamanoharan2024jumping}, with recent efforts extending this approach to ViTs \cite{lim2025sparseautoencodersrevealselective}.

However, post-hoc SAE methods introduce an important limitation. In a typical post-hoc setting, the SAE is trained on frozen representations from a pretrained model. Consequently, its latent features are learned independently of the downstream classification objective and may not adequately reflect the class-discriminative concepts required for the target task. Although such features may be sparse or visually coherent, they are not explicitly optimized to support the model's predictions.

To address this limitation, we propose SpIn-ViT, a machine learning framework that jointly trains a pretrained ViT and a modified SAE end-to-end. By directly coupling sparse representation learning with the classification objective, SpIn-ViT learns patch-level latent features that are both interpretable and aligned with the model's predictions.

We further assess whether SpIn-ViT's features support symbolic rule extraction. 
We binarize SpIn-ViT's latent activations and use FOLD-SE-M algorithm \cite{wang2022foldse} to induce a stratified Answer Set Program \cite{Baral}. Combined with the neural feature extractor, these rules form a neurosymbolic classifier. Compared with alternative SAE representations, SpIn-ViT yields more compact rules that preserve more classification accuracy, indicating greater suitability for symbolic composition and reasoning.

Across 9 benchmarks, SpIn-ViT achieves 88.30\% average classification accuracy, improving over the state-of-the-art post-hoc SAE by \textbf{8.84\%} while remaining competitive with vanilla ViT. It also attains nearly \textbf{4$\times$} higher AI-based and over \textbf{2$\times$} higher human interpretability scores. Its neurosymbolic models achieve \textbf{5.97\%} higher accuracy with \textbf{58.85\%} smaller rule sets than the post-hoc SAE baseline.

We summarize our contributions as follows:
\begin{enumerate}
    \item We propose SpIn-ViT, which jointly trains a pretrained ViT and modified SAE to align sparse patch-level representations with classification.

    \item We show that its latent features support compact rule extraction and accurate, interpretable neurosymbolic classifiers.

    \item We evaluate SpIn-ViT across 9 benchmark datasets using quantitative, AI-based and human evaluations, demonstrating improved interpretability and high accuracy.

\end{enumerate}

\begin{figure*}[t]
    \centering
    \includegraphics[height=5cm,width=1.0\textwidth,keepaspectratio]{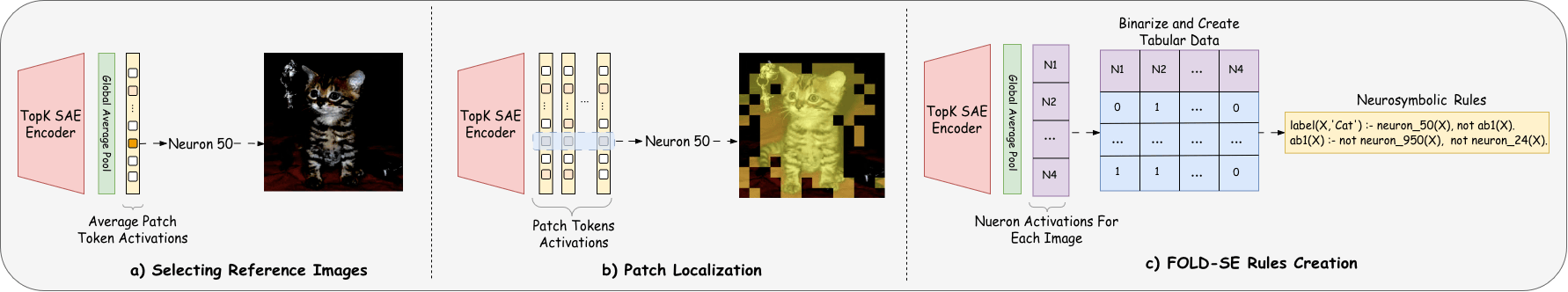}
    \caption{Interpretability analysis performed after training. (a) Reference images for a given neuron (e.g., Neuron 50) are selected by ranking [CLS] token activations from the trained TopK SAE encoder. (b) Patch-token activations for that neuron are then used to localize the specific image regions responsible for the activation. (c) Neuron activations across images are binarized into tabular data, over which FOLD-SE-M creates neurosymbolic rules for explainability of the TopK SAE encoder.}
    \label{fig:interp-analysis}
\end{figure*}

\section{Methodology}

In this section, we present the SpIn-ViT architecture and training objectives, quantitative and qualitative interpretability evaluations and neurosymbolic model creation. The framework is illustrated in Figure~\ref{fig:SpIn-ViT-Training}.

\medskip\noindent\textbf{Architecture Overview:}
\label{subsec:sae-architecture}
A Sparse Autoencoder (SAE) encodes dense representations into a higher-dimensional sparse latent space and reconstructs them through a decoder. Restricting each input to a few active neurons encourages distinct, potentially monosemantic features. In SpIn-ViT, the SAE processes ViT patch-token representations.

Unlike standard post-hoc SAEs trained unsupervised on frozen features, SpIn-ViT jointly fine-tunes the ViT and SAE with supervision, aligning sparse representations with the downstream task while improving interpretability without sacrificing classification performance.



SpIn-ViT builds on a trainable ViT encoder that maps an input image
$X \in \mathbb{R}^{C_{\text{in}} \times H \times W}$ to a sequence of patch-token representations
$\mathbf{z} = [\mathbf{z}_1, \mathbf{z}_2, \dots, \mathbf{z}_n] \in \mathbb{R}^{n \times d}$,
where $C_{\text{in}}$ denotes the number of input channels, $H$ and $W$ denote the image height and width, respectively, $n$ denotes the number of image patches, and $d=768$ is the embedding dimension. After the final transformer layer, the \texttt{[CLS]} token representation is passed through a linear classification head to produce the final class logits. The remaining patch-token representations are passed to the SAE module, whose encoder projects them into a sparse latent space and whose decoder reconstructs the original embeddings:
\begin{align}
\mathbf{h}
&=
\operatorname{TopK}
\left(
\mathbf{z}\mathbf{W}_{\text{enc}}
+
\mathbf{b}_{\text{enc}}
\right),
\label{eq:sae-encoder}
\\
\hat{\mathbf{z}}
&=
\mathbf{h}\mathbf{W}_{\text{dec}}
+
\mathbf{b}_{\text{dec}}.
\label{eq:sae-decoder}
\end{align}


The patch-token representations $\mathbf{z}$ are first projected into an $m$-dimensional latent pre-activation space using the SAE encoder weights $\mathbf{W}_{\text{enc}} \in \mathbb{R}^{d \times m}$ and bias $\mathbf{b}_{\text{enc}} \in \mathbb{R}^{m}$, where $m = 10 \times C$ and $C$ denotes the number of classes. SpIn-ViT then applies a TopK operator independently to each patch token. The operator retains the $k$ largest latent activations and sets all remaining entries to zero, producing the sparse latent representation $\mathbf{h} \in \mathbb{R}^{n \times m}$.

TopK fixes the number of active neurons per patch, preventing dense activations and restricting each region to a small set of salient features. It complements the $L_1$ penalty: TopK selects which neurons remain active, whereas $L_1$ regularizes their magnitudes. Together, they promote sparse, semantically disentangled representations.

The SAE decoder subsequently maps $\mathbf{h}$ back to the ViT embedding space using $\mathbf{W}_{\text{dec}} \in \mathbb{R}^{m \times d}$ and $\mathbf{b}_{\text{dec}} \in \mathbb{R}^{d}$, producing reconstructed patch-token representations $\hat{\mathbf{z}} \in \mathbb{R}^{n \times d}$ that approximate the original embeddings $\mathbf{z}$.

\medskip\noindent\textbf{Learning Objectives:}
\label{subsec:learn-obj}
SpIn-ViT is trained using four complementary learning objectives. The \textit{classification objective} promotes accurate class predictions and aligns the learned representations with the downstream task. The \textit{interpretability objective} encourages sparse latent activations while preserving the information contained in the original ViT patch embeddings. The \textit{intra-class objective} reduces variation among representations belonging to the same class, thereby improving class-level coherence. Finally, the \textit{inter-class objective} organizes the latent space into class-specific subspaces and encourages distinct, non-redundant features across classes.


\underline{\textit{Classification Objective:}}
The classification objective trains SpIn-ViT to produce accurate image-level predictions. Given a mini-batch of $B$ images and $C$ classes, we minimize the categorical cross-entropy loss
\begin{equation}
\mathcal{L}_{\text{cls}}
=
-\frac{1}{B}
\sum_{i=1}^{B}
\sum_{c=1}^{C}
y_{i,c}\log p_{i,c}
\label{eq:classification-loss}
\end{equation}
where $y_{i,c}$ is the ground-truth indicator specifying whether image $X_i$ belongs to class $c$. The predicted class-probability distribution is given by
$\mathbf{p}_i
=
\operatorname{softmax}\!\left(f_{\theta}(X_i)\right).$

$f_{\theta}$ denotes the SpIn-ViT classifier parameterized by the trainable parameters $\theta$, and $p_{i,c}$ denotes its predicted probability that image $X_i$ belongs to class $c$. The classifier applies a linear classification head to the final \texttt{[CLS]} token representation to produce the class logits. Minimizing $\mathcal{L}_{\text{cls}}$ aligns the learned representations with the downstream classification task.

\underline{\textit{Interpretability Objective:}}
The interpretability objective encourages the SAE to preserve information from the ViT patch-token representations while learning sparse and semantically disentangled latent features. Given a mini-batch of $B$ images, we define the SAE objective as
\begin{equation}
\mathcal{L}_{\text{SAE}}
=
\underbrace{
\frac{1}{B}
\sum_{i=1}^{B}
\left\|
\mathbf{z}_i-\hat{\mathbf{z}}_i
\right\|_{F}^{2}
}_{\mathcal{L}_{\text{recon}}}
+
\lambda
\underbrace{
\frac{1}{B}
\sum_{i=1}^{B}
\left\|
\mathbf{h}_i
\right\|_{1}
}_{\mathcal{L}_{\text{sparse}}}.
\label{eq:sae-loss}
\end{equation}

where $\mathbf{z}_i$ and $\hat{\mathbf{z}}_i$ denote the original and reconstructed patch-token representations for image $X_i$, respectively, and $\mathbf{h}_i$ denotes the corresponding sparse latent representation. The reconstruction term $\mathcal{L}_{\text{recon}}$ minimizes the discrepancy between the original and reconstructed ViT representations, thereby preserving information required to represent the input features. The regularization term $\mathcal{L}_{\text{sparse}}$ penalizes the magnitudes of the latent activations retained by the TopK operator. The coefficient $\lambda$ controls the balance between reconstruction fidelity and latent activation regularization. Together, TopK and the $L_1$ penalty encourage compact latent representations whose active neurons capture salient visual features.


\underline{\textit{Intra-Class Objective:}}
To improve class-level coherence in the reconstructed patch-token space, we introduce a center loss $\mathcal{L}_{\text{center}}$ that encourages samples from the same class to have similar representations:
\begin{equation}
    \mathcal{L}_{\text{center}}
    =
    \frac{1}{2B}
    \sum_{i=1}^{B}
    \left\|
    \mathbf{p}_i-\mathbf{c}_{y_i}
    \right\|_2^2,
\end{equation}
where $B$ denotes the mini-batch size, $\mathbf{p}_i \in \mathbb{R}^{d}$ is the mean-pooled SAE reconstruction for image $X_i$, $y_i$ is its ground-truth class label, and $\mathbf{c}_{y_i} \in \mathbb{R}^{d}$ denotes the center associated with class $y_i$. Specifically, $\mathbf{p}_i$ is obtained by averaging the reconstructed patch-token representations $\hat{\mathbf{z}}_i$ across the patch dimension.

Minimizing $\mathcal{L}_{\text{center}}$ reduces intra-class variation by drawing reconstructed representations toward their corresponding class centers. This encourages patch-level representations from the same class to form coherent clusters while preserving distinctions between classes. Because gradients propagate through the SAE and ViT encoder, the center loss also indirectly structures the latent representation $\mathbf{h}_i$ and aligns it with class-discriminative information.




\underline{\textit{Inter-Class Objective:}}
As the dimensionality of the SAE latent space increases, different latent neurons may learn redundant features. To structure the latent space and encourage feature diversity, we divide its $m$ dimensions into $C$ non-overlapping, class-indexed blocks, where $C$ denotes the number of classes and $r=m/C$ denotes the number of latent dimensions assigned to each block. For a mini-batch, we stack the patch-level latent representations across all images and patches to obtain
$\mathbf{H} \in \mathbb{R}^{Bn \times m}$, where $B$ is the mini-batch size and $n$ is the number of patches per image. We then partition 
$\mathbf{H}
=
\left[
\mathbf{S}^{(1)},
\mathbf{S}^{(2)},
\dots,
\mathbf{S}^{(C)}
\right],
$

where each block $\mathbf{S}^{(c)} \in \mathbb{R}^{Bn \times r}$ contains the latent activations associated with class-indexed subspace $c$.

To discourage redundant latent dimensions within each block, we apply the orthogonality regularizer
\begin{equation}
    \mathcal{L}_{\text{ortho}}
    =
    \frac{1}{C}
    \sum_{c=1}^{C}
    \left\|
    \left(\mathbf{S}^{(c)}\right)^{\top}
    \mathbf{S}^{(c)}
    -
    \mathbf{I}_{r}
    \right\|_{F}^{2},
\end{equation}
where $\mathbf{I}_{r} \in \mathbb{R}^{r \times r}$ is the identity matrix and $\|\cdot\|_{F}$ denotes the Frobenius norm. Minimizing $\mathcal{L}_{\text{ortho}}$ reduces correlations among latent dimensions within each class-indexed block, encouraging the neurons in each subspace to capture diverse and non-redundant visual features.

\underline{\textit{Total Learning Objective:}} The total learning objective integrates all components into a single unified loss function:

\[
\mathcal{L} = \alpha \mathcal{L}_{\text{cls}} + \beta \mathcal{L}_{\text{SAE}} + \delta \mathcal{L}_{\text{center}} + \gamma \mathcal{L}_{\text{ortho}}
\]

where $\alpha, \beta, \delta, \gamma \in [0, 1]$ are scalar weights. $\alpha$ controls the contribution of the classification loss, $\beta$ controls the contribution of the SAE loss, $\delta$ controls the contribution of the center loss, and $\gamma$ controls the contribution of the orthogonality regularizer. This balance ensures that SpIn-ViT jointly optimizes for prediction accuracy and interpretability.


\medskip\noindent\textbf{Quantitative Analysis Formulation:}
\label{subsec:quant-eval}
After training, we evaluate interpretability by selecting each neuron's strongest reference images, localizing its patch-level responses, and measuring whether the highlighted regions are prediction-relevant and semantically meaningful.

\underline{\textit{Selecting Reference Images:}}
As shown in Fig.~\ref{fig:interp-analysis}, we aggregate the SAE activations across all patches in each image to obtain a global activation vector
$\mathbf{p}_{\text{patch}}^{(i)}
=
\frac{1}{n}
\sum_{j=1}^{n}
\mathbf{h}_{i,j},
$

where $\mathbf{h}_{i,j} \in \mathbb{R}^{m}$ denotes the SAE latent representation associated with the $j$-th patch of image $X_i$, and $n$ denotes the number of image patches. Each entry of $\mathbf{p}_{\text{patch}}^{(i)}$ therefore represents the aggregate response of a latent neuron across the image.

For each latent neuron, we rank all images according to the corresponding values in $\mathbf{p}_{\text{patch}}^{(i)}$ and select the top-$K_{\text{ref}}$ images with the strongest responses as reference images. Examining the neuron's activations across the patches of these reference images reveals the visual patterns that most consistently evoke its response.


\underline{\textit{Patch Segmentation:}}
After selecting the reference images, we localize the image regions associated with each SAE neuron by examining its activation across individual patches, as shown in Fig.~\ref{fig:interp-analysis}. For an image $X_i$, let $\mathbf{h}_i \in \mathbb{R}^{n \times m}$ denote its SAE latent representation, where $n$ is the number of image patches and $m$ is the number of SAE neurons. The activation of neuron $q$ for patch $j$ is denoted by $h_{i,j,q}$.

During inference, we apply a Sigmoid function to the latent activations to map them to the interval $[0,1]$. We then threshold the normalized activations at $\tau=0.5$ to obtain a binary patch-activation mask:
\begin{equation}
\tilde{h}_{i,j,q}
=
\sigma\!\left(h_{i,j,q}\right),
\qquad
A_{i,j}^{(q)}
=
\begin{cases}
1, & \text{if } \tilde{h}_{i,j,q} > 0.5, \\
0, & \text{otherwise},
\end{cases}
\end{equation}
where $\sigma(\cdot)$ denotes the Sigmoid function. The corresponding segmented patch is given by
$S_{i,j}^{(q)}
=
A_{i,j}^{(q)} \odot X_{i,j}$, where $X_{i,j}$ denotes the $j$-th image patch of $X_i$, and $\odot$ denotes element-wise multiplication. The binary mask $\mathbf{A}_i^{(q)}$ identifies the patches associated with neuron $q$, while $\mathbf{S}_i^{(q)}$ retains the corresponding image regions. Reassembling the masked patches in their original spatial arrangement produces a patch-wise segmentation map that visualizes the regions associated with the neuron's response.

\underline{\textit{Insertion and Deletion:}}
Insertion and Deletion metrics evaluate whether the neuron's responses are spatially grounded in image regions that influence the model's prediction \cite{petsiuk2018rise}. Deletion measures the decrease in model confidence as the most relevant image patches are progressively removed, whereas insertion measures the increase in confidence as those patches are progressively revealed from a blank image. We rank the patches according to their continuous neuron activation scores $\tilde{h}_{i,j,q}$ and use the corresponding patch-wise segmentation map $\mathbf{S}_{i}^{(q)}$ to perform the insertion and deletion procedures. A lower deletion score and a higher insertion score indicate that the neuron activations identify image regions that are more faithful to the model's prediction.


\begin{figure*}[t]
    \centering
    \includegraphics[height=6cm,width=1.0\textwidth]{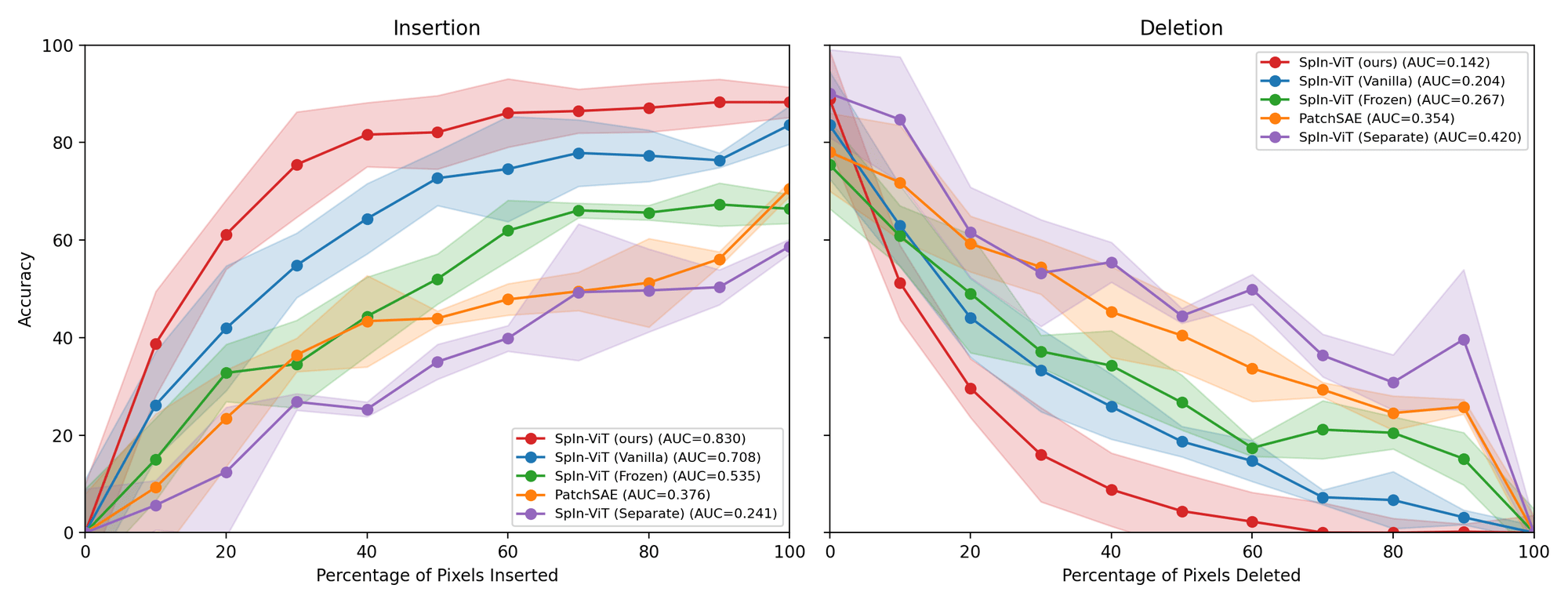}
    \caption{Insertion and Deletion Plots of each model for average of all datasets}
    \label{fig:insert-delete}
\label{fig:foldse_rules}
\end{figure*}

\begin{table*}[h]
\centering
\small
\caption{Classification accuracy (\%) across nine benchmark datasets (mean $\pm$ std). Best results per dataset are highlighted in \textbf{bold}.}
\label{tab:accuracy}
\setlength{\tabcolsep}{4pt}
\begin{tabular}{lcccccc}
\toprule
\textbf{Dataset} & \textbf{PatchSAE} & \textbf{Vanilla ViT} & \textbf{SpIn-ViT (Separate)} & \textbf{SpIn-ViT (Frozen)} & \textbf{SpIn-ViT (Vanilla)} & \textbf{SpIn-ViT (Ours)} \\
\midrule
Flowers102       & 94.99 $\pm$ 0.42        & 98.54 $\pm$ 0.21        & 97.94 $\pm$ 0.38        & 98.71 $\pm$ 0.29        & 98.90 $\pm$ 0.19        & \textbf{99.88 $\pm$ 0.28} \\
Caltech101       & 95.10 $\pm$ 0.31        & 96.80 $\pm$ 0.58        & 95.70 $\pm$ 0.24        & 93.47 $\pm$ 0.62        & 96.81 $\pm$ 0.23        & \textbf{97.02 $\pm$ 0.18} \\
Stanford Cars    & 75.71 $\pm$ 1.53        & 82.37 $\pm$ 1.24        & 82.57 $\pm$ 1.39        & 66.46 $\pm$ 1.64        & 82.14 $\pm$ 1.21        & \textbf{84.61 $\pm$ 1.02} \\
FGVC             & 55.43 $\pm$ 2.18        & \textbf{71.11 $\pm$ 1.67} & 68.32 $\pm$ 2.01        & 70.12 $\pm$ 1.75        & 67.63 $\pm$ 1.82        & 68.43 $\pm$ 1.79        \\
EuroSAT          & 78.58 $\pm$ 1.08        & 97.41 $\pm$ 0.22        & 97.31 $\pm$ 0.21        & 62.42 $\pm$ 1.88        & \textbf{98.62 $\pm$ 0.18} & 98.58 $\pm$ 0.21        \\
DTD              & 76.42 $\pm$ 1.32        & 80.90 $\pm$ 0.93        & 80.78 $\pm$ 0.98        & 53.44 $\pm$ 2.31        & 82.59 $\pm$ 1.11        & \textbf{84.42 $\pm$ 1.28} \\
Sun397           & 77.14 $\pm$ 1.27        & 76.54 $\pm$ 1.12        & 76.43 $\pm$ 1.22        & 72.27 $\pm$ 1.89        & 78.01 $\pm$ 1.04        & \textbf{80.00 $\pm$ 1.38} \\
Food101          & 84.81 $\pm$ 1.14        & 85.08 $\pm$ 1.25        & 85.13 $\pm$ 0.82        & 86.65 $\pm$ 1.12        & 85.55 $\pm$ 1.22        & \textbf{86.81 $\pm$ 1.11} \\
OxfordPet        & 91.96 $\pm$ 0.65        & 90.32 $\pm$ 0.96        & 90.22 $\pm$ 0.85        & 92.48 $\pm$ 1.02        & 92.09 $\pm$ 0.30        & \textbf{94.96 $\pm$ 0.85} \\
\midrule
\textbf{Average} & 81.13 $\pm$ 1.10        & 86.56 $\pm$ 0.91        & 86.04 $\pm$ 0.90        & 77.34 $\pm$ 1.39        & 86.93 $\pm$ 0.81        & \textbf{88.30 $\pm$ 0.90} \\
\bottomrule
\end{tabular}
\end{table*}

\begin{table*}[t]
\centering
\caption{AI interpretability evaluations across nine image classification benchmarks (normalized to a 1--5 scale where 1.00 = 0.0 mIoU and 5.00 = 1.0 mIoU).}
\begin{tabular}{lccccc}
\toprule
\textbf{Dataset} & \textbf{PatchSAE} & \textbf{SpIn-ViT (Frozen)} & \textbf{SpIn-ViT (Separate)} & \textbf{SpIn-ViT (Vanilla)} & \textbf{SpIn-ViT (Ours)} \\
\midrule
Flowers102    & $1.32 \pm 0.18$ & $1.60 \pm 0.24$ & $1.08 \pm 0.12$ & $3.00 \pm 0.32$ & $\mathbf{3.28 \pm 0.35}$ \\
Caltech101    & $1.32 \pm 0.22$ & $1.56 \pm 0.22$ & $1.08 \pm 0.11$ & $2.96 \pm 0.30$ & $\mathbf{3.24 \pm 0.33}$ \\
Stanford Cars & $1.24 \pm 0.15$ & $1.40 \pm 0.18$ & $1.04 \pm 0.10$ & $2.76 \pm 0.25$ & $\mathbf{3.00 \pm 0.28}$ \\
FGVC          & $1.16 \pm 0.12$ & $1.28 \pm 0.15$ & $1.04 \pm 0.10$ & $2.56 \pm 0.20$ & $\mathbf{2.72 \pm 0.22}$ \\
EuroSAT       & $1.36 \pm 0.25$ & $1.64 \pm 0.28$ & $1.08 \pm 0.12$ & $3.04 \pm 0.35$ & $\mathbf{3.28 \pm 0.36}$ \\
DTD           & $1.24 \pm 0.16$ & $1.44 \pm 0.19$ & $1.04 \pm 0.10$ & $2.76 \pm 0.26$ & $\mathbf{3.00 \pm 0.27}$ \\
Sun397        & $1.20 \pm 0.14$ & $1.40 \pm 0.17$ & $1.04 \pm 0.10$ & $2.68 \pm 0.22$ & $\mathbf{2.92 \pm 0.24}$ \\
Food101       & $1.28 \pm 0.19$ & $1.48 \pm 0.21$ & $1.08 \pm 0.11$ & $2.84 \pm 0.28$ & $\mathbf{3.08 \pm 0.31}$ \\
OxfordPet     & $1.32 \pm 0.21$ & $1.56 \pm 0.23$ & $1.08 \pm 0.12$ & $2.92 \pm 0.29$ & $\mathbf{3.20 \pm 0.32}$ \\
\midrule
\textbf{Average} & $1.27 \pm 0.18$ & $1.48 \pm 0.21$ & $1.06 \pm 0.11$ & $2.84 \pm 0.27$ & $\mathbf{3.08 \pm 0.30}$ \\
\bottomrule
\end{tabular}
\label{tab:ai-evals}
\end{table*}

\medskip\noindent\textbf{Qualitative Analysis Formulation:}
\label{subsec:qual-eval}
Quantitative metrics do not establish whether explanations correspond to human-meaningful visual concepts. We therefore complement them with a human study of SAE-highlighted regions and an AI-based evaluation across all nine datasets.

\underline{\textit{Human Evaluation:}}
We selected 50 images each from the datasets with the highest and lowest SpIn-ViT classification accuracy, yielding 100 images evaluated by 12 participants. For each image, participants rated whether the patch-wise segmentation map $\mathbf{S}_{i}^{(q)}$ from the most active SAE neuron $q$ highlighted a coherent, recognizable visual concept. This evaluates perceptual interpretability under both strong and weak classification performance.

\underline{\textit{AI-Based Evaluation:}}
To extend beyond the limited human study, we conduct an automated evaluation across all nine datasets. For each image, SAM3 \cite{carion2025sam3segmentconcepts} generates a reference mask from a text prompt describing the target concept. We compute the mean Intersection over Union (mIoU) between this mask and each method's patch-wise segmentation map $\mathbf{S}_{i}^{(q)}$, where higher mIoU indicates stronger spatial agreement. This evaluation provides broader coverage and tests whether the trends observed in the human study generalize across datasets. Additional details are provided in the supplementary material.

\section{Neurosymbolic Model Creation}
\label{sec:nesy-model}

Prior work combines neural feature extractors with symbolic rules induced from internal representations, yielding human-readable predicates grounded in neurons \cite{eric,nesyfold,padalkar2025symbolic}. Although these neurosymbolic models often trail the original network in accuracy, sparse, distinct, and class-discriminative features can reduce both this gap and rule-set size \cite{padalkar2025symbolic}.

Because SpIn-ViT learns sparse, semantically coherent SAE neurons aligned with classification, we evaluate their suitability for symbolic rule extraction. Although compatible with other induction algorithms, we use FOLD-SE-M \cite{wang2022foldse} for consistency with prior frameworks.



FOLD-SE-M is a rule-learning algorithm that induces classification rules from labeled tabular data as a stratified Answer Set Program (ASP); Fig.~\ref{fig:interp-analysis} shows an example. It learns default rules covering positive examples while limiting false positives, then recursively learns exceptions by swapping positive and negative examples. Exceptions are represented by abnormality predicates such as \texttt{ab\textit{x}}, where \texttt{\textit{x}} is a unique identifier.

Its two main hyperparameters control the accuracy--complexity trade-off: \texttt{ratio} limits the ratio of false positives to true positives covered by a rule's default component, while \texttt{tail} specifies the minimum number of training examples a rule must cover.



\textbf{Rule Extraction from SpIn-ViT:}
For each input image $X_i$, SpIn-ViT produces an SAE representation
$\mathbf{h}_i \in \mathbb{R}^{n \times m}$, where $n$ denotes the number of image patches and $m$ denotes the number of SAE neurons. We aggregate the activations across the patch dimension to obtain one feature vector per image: $\mathbf{p}_i
    =
    \frac{1}{n}
    \sum_{j=1}^{n}
    \mathbf{h}_{i,j}$, where $\mathbf{p}_i \in \mathbb{R}^{m}$ contains the mean activation of each SAE neuron for image $X_i$. We then apply the inference-time Sigmoid transformation and threshold the resulting activations to obtain binary values: $    a_{i,q}
    =
    \begin{cases}
        1, & \text{if } \sigma(p_{i,q}) > 0.5, \\
        0, & \text{otherwise},
    \end{cases}$ 
where $a_{i,q}$ denotes the truth value of the predicate associated with SAE neuron $q$ for image $X_i$.

Each image is represented as a row in a tabular dataset, with the binary SAE neuron activations used as input attributes and the ground-truth class used as the target label. FOLD-SE-M is then applied to this representation to generate rules that express class predictions in terms of active SAE neurons. For example, a rule may predict a class when a particular combination of neuron predicates is true, unless an exception represented by a different combination of predicates is satisfied.

During inference, the ViT and SAE components first produce the neuron activations for an input image. These activations determine the truth values of their corresponding predicates, after which the extracted ASP rule set determines the predicted class. The trained neural feature extractor up to the SAE latent representation, together with the induced rule set, constitutes the complete neurosymbolic model. Because each predicate is grounded in an SAE neuron whose visual semantics can be examined through its patch-activation maps, the resulting rules provide logical explanations composed of visually interpretable neural features.

\begin{table}[t]
\centering
\caption{Human-rated interpretability of neuron activations (1--5, higher is better) on Flowers102 and FGVC.}
\small
\setlength{\tabcolsep}{2pt}
\begin{tabular}{lccc}
\toprule
\textbf{Method} & \textbf{Flowers102} & \textbf{FGVC} & \textbf{Average} \\
\midrule
PatchSAE            & $1.78 \pm 0.15$ & $1.92 \pm 0.17$ & $1.85 \pm 0.16$ \\
SpIn-ViT (Frozen)   & $2.45 \pm 0.18$ & $2.34 \pm 0.16$ & $2.40 \pm 0.17$ \\
SpIn-ViT (Separate) & $1.45 \pm 0.11$ & $1.32 \pm 0.12$ & $1.39 \pm 0.11$ \\
SpIn-ViT (Vanilla)  & $3.43 \pm 0.22$ & $3.22 \pm 0.24$ & $3.33 \pm 0.23$ \\
\textbf{SpIn-ViT (ours)} & $\mathbf{4.20 \pm 0.28}$ & $\mathbf{3.99 \pm 0.26}$ & $\mathbf{4.10 \pm 0.27}$ \\
\bottomrule
\end{tabular}
\label{tab:human-evals}
\end{table}

\section{Experimental Setup and Evaluations}
In this section, we evaluate SpIn-ViT and baselines on classification performance, AI-based and human interpretability, and neurosymbolic model performance.

\medskip\noindent\textbf{Implementation Details:}\\
\indent \underline{\textit{Datasets.}} We utilized 9 widely used benchmark datasets: Flowers102 \cite{nilsback2008automated}, Caltech101 \cite{4756141}, StanfordCars \cite{6755945}, FGVC Aircraft \cite{maji2013fine}, EuroSAT \cite{helber2019eurosat}, DTD \cite{cimpoi2014describing}, SUN397 \cite{xiao2010sun}, Food101 \cite{bossard2014food}, and OxfordPets \cite{parkhi2012cats}.

\underline{\textit{Training Configurations.}} Following \cite{lim2025patchsae}, we use a 5:1 stratified train/test split per class and adopt the same default hyperparameters for SpIn-ViT and all other baselines. All models are also trained with the AdamW optimizer \cite{loshchilov2018decoupled} (learning rate $1\times10^{-5}$, weight decay $0.01$, $\beta_1=0.9$, $\beta_2=0.999$), and a batch size of $16$ for each dataset.

\underline{\textit{Baselines.}}
We compare our proposed methods against several baselines used across all experiments, covering a range of architectural design choices and previous popular methods. Vanilla ViT \cite{dosovitskiy2021animage} is used only as a classification baseline with no SAE. PatchSAE \cite{lim2025patchsae} is the most recent interpretability-based baseline, applying a frozen, post-hoc SAE to pretrained CLIP-based ViT \cite{radford2021clip} features. SpIn-ViT (Separate) is a variant of our proposed method which first fine-tunes the vanilla ViT on the classification task alone, then fine-tunes on our proposed TopK SAE on the resulting downstream features. SpIn-ViT (Frozen) has a similar fine-tuning strategy of that of PatchSAE. We use our own TopK SAE architecture in place of PatchSAE's, while keeping the ViT backbone frozen. SpIn-ViT (Vanilla) is our full method without the TopK sparsity constraint. SpIn-ViT (Ours) is our proposed method, adding TopK sparsity along with center loss and orthogonality objectives to encourage class-aligned, monosemantic features.
\begin{table*}[t]
\centering
\caption{Accuracy (\%) and number of active rules ($\pm$ standard deviation) across nine image classification benchmarks.}
\label{tab:accuracy_and_rules}
\resizebox{\textwidth}{!}{%
\begin{tabular}{lcccccccccc}
\toprule
& \multicolumn{2}{c}{\textbf{PatchSAE}} & \multicolumn{2}{c}{\textbf{SpIn-ViT (Frozen)}} & \multicolumn{2}{c}{\textbf{SpIn-ViT (Separate)}} & \multicolumn{2}{c}{\textbf{SpIn-ViT (Vanilla)}} & \multicolumn{2}{c}{\textbf{SpIn-ViT (Ours)}} \\
\cmidrule(lr){2-3} \cmidrule(lr){4-5} \cmidrule(lr){6-7} \cmidrule(lr){8-9} \cmidrule(lr){10-11}
\textbf{Dataset} & \textbf{Acc ($\uparrow$)} & \textbf{Rules ($\downarrow$)} & \textbf{Acc ($\uparrow$)} & \textbf{Rules ($\downarrow$)} & \textbf{Acc ($\uparrow$)} & \textbf{Rules ($\downarrow$)} & \textbf{Acc ($\uparrow$)} & \textbf{Rules ($\downarrow$)} & \textbf{Acc ($\uparrow$)} & \textbf{Rules ($\downarrow$)} \\
\midrule
Flowers102     & $90.91 \pm 0.98$ & $20 \pm 3.4$ & $94.24 \pm 0.35$ & $14 \pm 1.2$ & $92.21 \pm 0.48$ & $24 \pm 2.8$ & $95.04 \pm 0.31$ & $13 \pm 2.5$ & $\mathbf{96.48 \pm 0.97}$ & $\mathbf{10 \pm 0.8}$ \\
Caltech101     & $89.62 \pm 0.38$ & $12 \pm 1.8$ & $93.56 \pm 0.16$ & $23 \pm 4.5$ & $90.97 \pm 1.13$ & $16 \pm 2.1$ & $94.38 \pm 0.43$ & $7 \pm 2.7$ & $\mathbf{95.16 \pm 0.78}$ & $\mathbf{5 \pm 0.9}$ \\
Stanford Cars  & $84.36 \pm 0.54$ & $48 \pm 5.8$ & $86.32 \pm 0.17$ & $38 \pm 3.2$ & $84.82 \pm 0.48$ & $24 \pm 4.6$ & $87.91 \pm 0.56$ & $30 \pm 2.9$ & $\mathbf{88.78 \pm 0.33}$ & $\mathbf{23 \pm 1.6}$ \\
FGVC           & $75.33 \pm 0.55$ & $106 \pm 6.2$ & $78.68 \pm 1.12$ & $67 \pm 4.8$ & $78.29 \pm 0.67$ & $60 \pm 3.1$ & $79.44 \pm 0.52$ & $65 \pm 2.4$ & $\mathbf{81.24 \pm 0.96}$ & $\mathbf{47 \pm 1.1}$ \\
EuroSAT        & $94.16 \pm 0.87$ & $60 \pm 4.9$ & $95.94 \pm 0.59$ & $40 \pm 3.7$ & $95.18 \pm 0.97$ & $34 \pm 2.3$ & $96.59 \pm 0.78$ & $40 \pm 1.4$ & $\mathbf{97.89 \pm 1.13}$ & $\mathbf{29 \pm 0.7}$ \\
DTD            & $79.37 \pm 0.97$ & $136 \pm 6.5$ & $83.46 \pm 0.93$ & $88 \pm 3.4$ & $82.23 \pm 0.81$ & $50 \pm 2.2$ & $83.34 \pm 1.14$ & $\mathbf{43 \pm 2.8}$ & $\mathbf{84.96 \pm 0.83}$ & $55 \pm 1.3$ \\
Sun397         & $74.12 \pm 0.72$ & $162 \pm 4.2$ & $78.43 \pm 0.86$ & $93 \pm 3.1$ & $75.82 \pm 0.74$ & $64 \pm 2.7$ & $78.77 \pm 0.43$ & $76 \pm 2.1$ & $\mathbf{79.61 \pm 0.46}$ & $\mathbf{63 \pm 1.0}$ \\
Food101        & $81.56 \pm 1.05$ & $161 \pm 4.1$ & $84.17 \pm 1.16$ & $93 \pm 2.9$ & $83.12 \pm 1.16$ & $101 \pm 3.8$ & $85.38 \pm 0.43$ & $74 \pm 2.6$ & $\mathbf{86.36 \pm 0.37}$ & $\mathbf{58 \pm 1.2}$ \\
OxfordPet      & $89.13 \pm 1.14$ & $80 \pm 2.4$ & $91.97 \pm 0.93$ & $53 \pm 3.1$ & $90.56 \pm 0.92$ & $38 \pm 1.9$ & $92.64 \pm 0.78$ & $48 \pm 2.7$ & $\mathbf{93.56 \pm 0.98}$ & $\mathbf{33 \pm 2.3}$ \\
\midrule
\textbf{Average} & $84.28 \pm 0.80$ & $94 \pm 4.4$ & $87.46 \pm 0.70$ & $56 \pm 3.3$ & $86.02 \pm 0.82$ & $46 \pm 2.7$ & $88.61 \pm 0.60$ & $44 \pm 2.4$ & $\mathbf{89.34 \pm 0.76}$ & $\mathbf{42 \pm 1.2}$ \\
\bottomrule
\label{tab:rules-creation}
\end{tabular}%
}
\end{table*}
\vspace{-8pt}

\medskip\noindent\textbf{Classification Results:} SpIn-ViT demonstrates strong and consistent classification performance across nine benchmark datasets as summarized in Table~\ref{tab:accuracy}. The SpIn-ViT (Frozen) variant degrades substantially, most severely on FGVC, EuroSAT, and DTD, confirming that post-hoc application of a frozen SAE on fixed ViT features introduces significant information loss. SpIn-ViT (Separate) recovers much of this degradation but still lags behind SpIn-ViT on most datasets, notably FGVC and Stanford Cars, demonstrating that unfreezing the SAE alone, without our structured center loss and orthogonality objectives, is insufficient for maintaining both classification performance and interpretability.

\medskip\noindent\textbf{Insertion and Deletion Analysis:}
Figure~\ref{fig:insert-delete} reports insertion and deletion curves complementing classifcation results. In the insertion test, accuracy should rise quickly as the highest-attributed pixels are added back. In the deletion test, accuracy should drop quickly as those same pixels are removed. 

SpIn-ViT (Ours) achieves the highest insertion AUC (0.830) and the lowest deletion AUC (0.142), rising fastest in insertion and falling fastest in deletion, indicating its neuron activations point to the pixels the model actually relies on for its predictions. SpIn-ViT (Vanilla) and SpIn-ViT (Frozen) follow the same ordering but with weaker separation between the two curves, while PatchSAE and SpIn-ViT (Separate) show the flattest insertion curves and slowest deletion drop-off, reflecting more diffuse, less faithful attributions. This ordering matches our earlier rule-creation and interpretability results, reinforcing that joint training produces activations more tightly aligned with class-discriminative image regions.

\medskip\noindent\textbf{Human Evaluation Results:}
Table~\ref{tab:human-evals} reports ratings from 12 participants on how well activated patches correspond to coherent visual concepts in Flowers102 and FGVC, using a 1--5 scale. SpIn-ViT (Ours) achieves the highest average score of 4.10, followed by SpIn-ViT (Vanilla) at 3.33. SpIn-ViT (Frozen), PatchSAE, and SpIn-ViT (Separate) score 2.40, 1.85, and 1.39, respectively. These results indicate that joint training improves perceptual alignment and that TopK sparsity further encourages activations corresponding to semantically meaningful image regions.


\medskip\noindent\textbf{AI Evaluation:}
\label{sec:ai_eval}
We compute mIoU between neuron-highlighted image regions and SAM3-generated concept masks \cite{carion2025sam3segmentconcepts}, then normalize scores to a 1--5 scale for comparison with human ratings. SpIn-ViT (Ours) ranks highest across all nine datasets, followed by SpIn-ViT (Vanilla), SpIn-ViT (Frozen), PatchSAE, and SpIn-ViT (Separate). This ordering persists on challenging datasets such as FGVC and SUN397 and closely matches the human evaluations, indicating agreement between automated and human judgments.



\medskip\noindent\textbf{Neurosymbolic Model Analysis:}
\label{sec:rules_creation}
Table~\ref{tab:rules-creation} reports classification accuracy alongside the size of the symbolic rule-set extracted for each method. High accuracy with a small rule-set size is desirable, where rule-set size is the total number of predicates in the rule-set body; smaller rule-sets correlate with higher interpretability \cite{rulesetinterpretability}. All hyperparameters are in the supplementary material.

SpIn-ViT (Ours) achieves the highest accuracy on all nine datasets and the fewest rules on eight of nine, trailing only SpIn-ViT (Frozen) on DTD despite a 2.4 percent accuracy gain. This pattern is clearest on Caltech101 and Flowers102, where we reach top accuracy with only 5 and 10 rules, well below PatchSAE's 12 and 20 despite PatchSAE's lower accuracy, evidence that PatchSAE's features are more entangled and require extra rules to compensate. The remaining variants help explain why: SpIn-ViT (Frozen) often yields the fewest rules but the lowest accuracy, since decoupling the SAE from the classification objective produces a constrained but less discriminative feature space, while Separate and Vanilla trade off between the two. Only joint optimization achieves both together, confirming that features shaped directly by the classification signal are more distinguishable and more compactly rule-summarizable. The same pattern holds even on harder datasets like FGVC and Sun397, where other methods need far more rules while Ours remains both the most compact and most accurate.

\section{Related Works}
\textbf{SAE for Mechanistic Interpretability.} SAEs are widely used to disentangle model representations into monosemantic features, most notably in LLMs, where they are attached to each transformer layer to extract increasingly fine-grained features. SAE-derived latent directions have also enabled downstream steering, such as style modulation \cite{konen2024stylevectors}, refusal behavior \cite{arditi2024refusaldirection}, and alignment-oriented activation editing \cite{kong2024control, qiu2024spectral}, showing that directions like refusal, bias, or truthfulness can be reliably manipulated to alter outputs. SAEs have similarly extended to vision: applications to diffusion models uncover factors governing denoising trajectories and spatial layout \cite{liu2024saediffusion}, sparse decompositions of generative models isolate concepts like texture and shape \cite{bai2024conceptsae}, and in ViTs, SAEs recover token level features aligned with meaningful visual attributes \cite{jing2024saevit}. Together, these results show that the monosemantic structure captured by SAEs generalizes beyond language, offering a general mechanism for interpreting high-dimensional representations across vision and generative models.

\textbf{Neurosymbolic AI for Vision Models.} Neurosymbolic approaches bind learned image features to explicit, human-auditable rules rather than leaving them as opaque representations. In CNNs, this has taken the form of extracting logic-based rule-sets from convolutional filter image features, such as NeSyFOLD \cite{nesyfold}, which uses the FOLD-SE-M algorithm \cite{wang2022foldse} to convert binarized kernel image features into a rule-set, and follow-up work improving the accuracy and compactness of these extracted rules \cite{padalkar2025classspecific}. Extending this to ViTs is more difficult due to their lack of modular concept detectors, though recent work has introduced a sparse concept layer to enable rule extraction from attention-weighted patch image features \cite{padalkar2025symbolic}. We build on this line of work, using FOLD-SE-M to extract symbolic rules directly from SpIn-ViT's jointly trained latent image features.

\section{Conclusion}
We introduced SpIn-ViT, which integrates sparse autoencoders into the classification objective to learn interpretable patch-level representations while maintaining competitive accuracy. Across nine benchmarks, it outperforms post-hoc SAE baselines on quantitative, AI-based, and human interpretability evaluations while matching or exceeding vanilla ViT performance. Its latent neurons also enable FOLD-SE-M to construct more accurate neurosymbolic models with smaller rule sets than post-hoc SAE representations. Overall, we demonstrate that high levels of interpretability, accuracy, and neurosymbolic reasoning can be achieved together.

\bibliography{aaai2027}

\clearpage
\setcounter{secnumdepth}{2}
\appendix
\section{Appendix}
\subsection{Training and Evaluation Protocol}
\label{sec:protocol}

\textbf{SpIn-ViT Hyperparameters.} The overall training objective combines a classification loss $\mathcal{L}_{cls}$, an SAE reconstruction loss $\mathcal{L}_{SAE}$, a center loss $\mathcal{L}_{center}$, and an orthogonality loss $\mathcal{L}_{ortho}$, weighted by coefficients $\alpha$, $\beta$, $\delta$, and $\gamma$ respectively. We set $\alpha = 1$ and $\beta = 1$ to weight classification and reconstruction losses equally, while the sparsity term within $\mathcal{L}_{SAE}$ is set to $\lambda = 5\text{e-}4$, following prior SAE sparsity conventions~\cite{bricken2023towards}. The learning objective weights $\delta$ and $\gamma$ were selected through a manual grid search over $\delta \in \{0.1, 0.3, 0.5, 1.0\}$ and $\gamma \in \{0.5, 1.0, 2.0, 4.0\}$, with final values chosen based on validation accuracy on a held-out split of each dataset. We set $\delta = 0.3$ to pull class-conditional logits toward their respective class centers, tightening intra-class clustering. We set $\gamma = 2.0$ to enforce separation between sparse activation directions across classes, discouraging redundant or overlapping patch-level features. Random seed 42 was used for all runs.

\textbf{FOLD-SE Hyperparameters.} For rule extraction via FOLD-SE-M, the exception ratio was set to $ratio = 1.0$ and the covering limit to $tail = 5\text{e-}4$.

\textbf{Computing infrastructure.} All experiments were run on a single NVIDIA RTX 4090 GPU with 24GB VRAM.

\subsection{Further Insertion-Deletion Evaluations and Rule Extraction}
\label{sec:appendix_id_rules}

In this section, we provide additional evidence beyond the averaged results reported in the main paper. This includes per-dataset insertion and deletion results and the correspondence between extracted symbolic rules and SAE neuron activations.

\subsubsection{Insertion Results}
\label{sec:insertion}
\begin{figure*}[p]
   \includegraphics[width=1.0\textwidth]{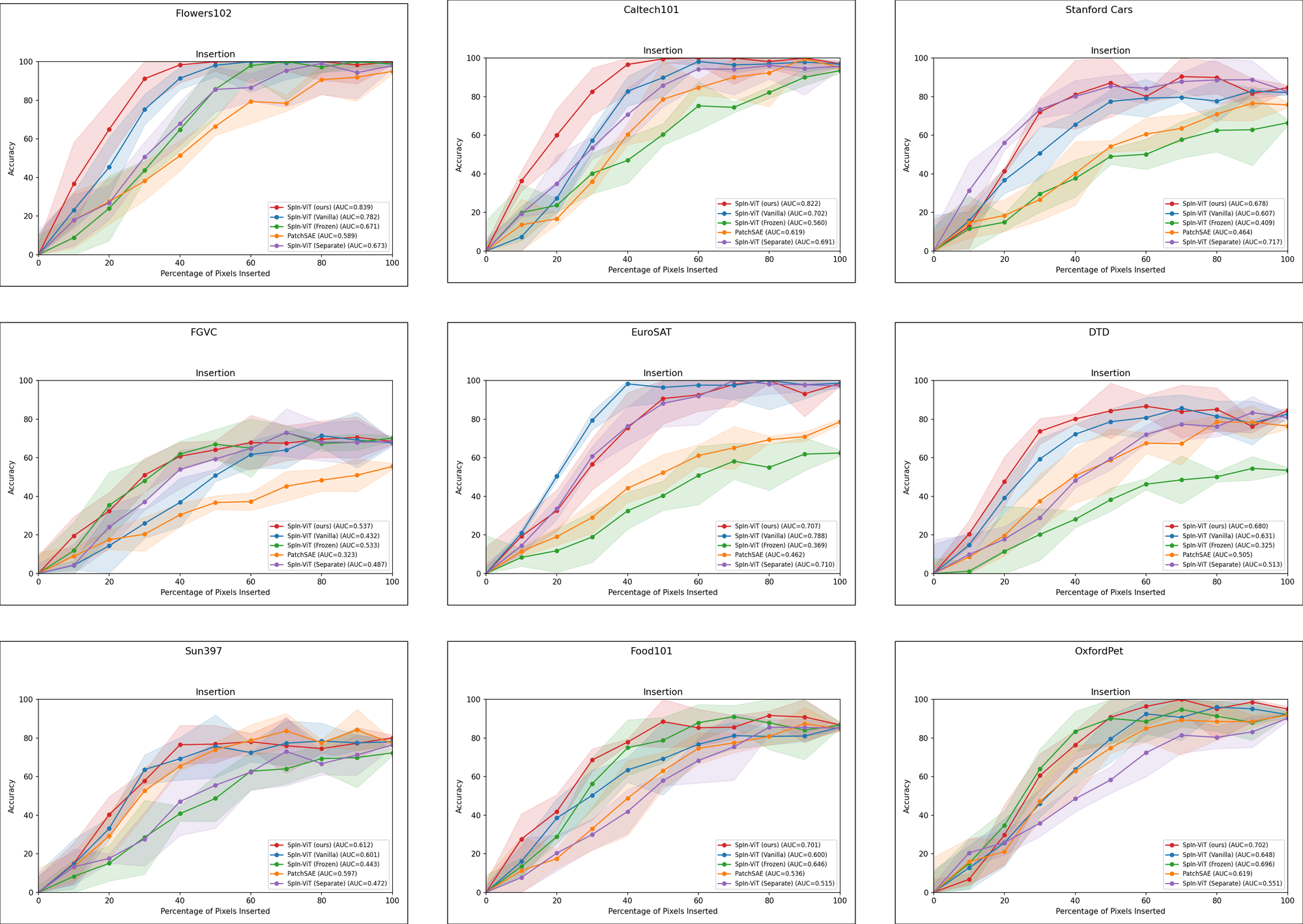}
   \caption{Insertion plots for all nine benchmark datasets. For each dataset, pixels are progressively inserted into a blank baseline image in order of attribution ranking (most salient first, per each model's attribution map), and classification accuracy is plotted as a function of the percentage of pixels inserted. Higher curves and larger AUC indicate more faithful attributions.}
   \label{fig:insert}
\end{figure*}

Figure~\ref{fig:insert} shows per-dataset insertion results. SpIn-ViT (ours) achieves the highest or near-highest AUC on most benchmarks, notably Flowers102~\cite{nilsback2008automated}, Caltech101~\cite{4756141}, EuroSAT~\cite{helber2019eurosat}, and DTD~\cite{cimpoi2014describing}, where accuracy rises sharply within the first 20--30\% of inserted pixels and plateaus before the baselines. This indicates our most salient patches are also the most sufficient for correct classification. On FGVC~\cite{maji2013fine}, Sun397~\cite{xiao2010sun}, and Food101~\cite{bossard2014food}, the gap narrows, suggesting attribution sharpness is harder when class-distinguishing evidence is more diffuse. PatchSAE~\cite{lim2025patchsae} and SpIn-ViT (Separate) insert more slowly, reflecting less class-aligned patches relative to our jointly trained representations.

\subsubsection{Deletion Results}
\label{sec:deletion}
\begin{figure*}[p]
   \includegraphics[width=1.0\textwidth]{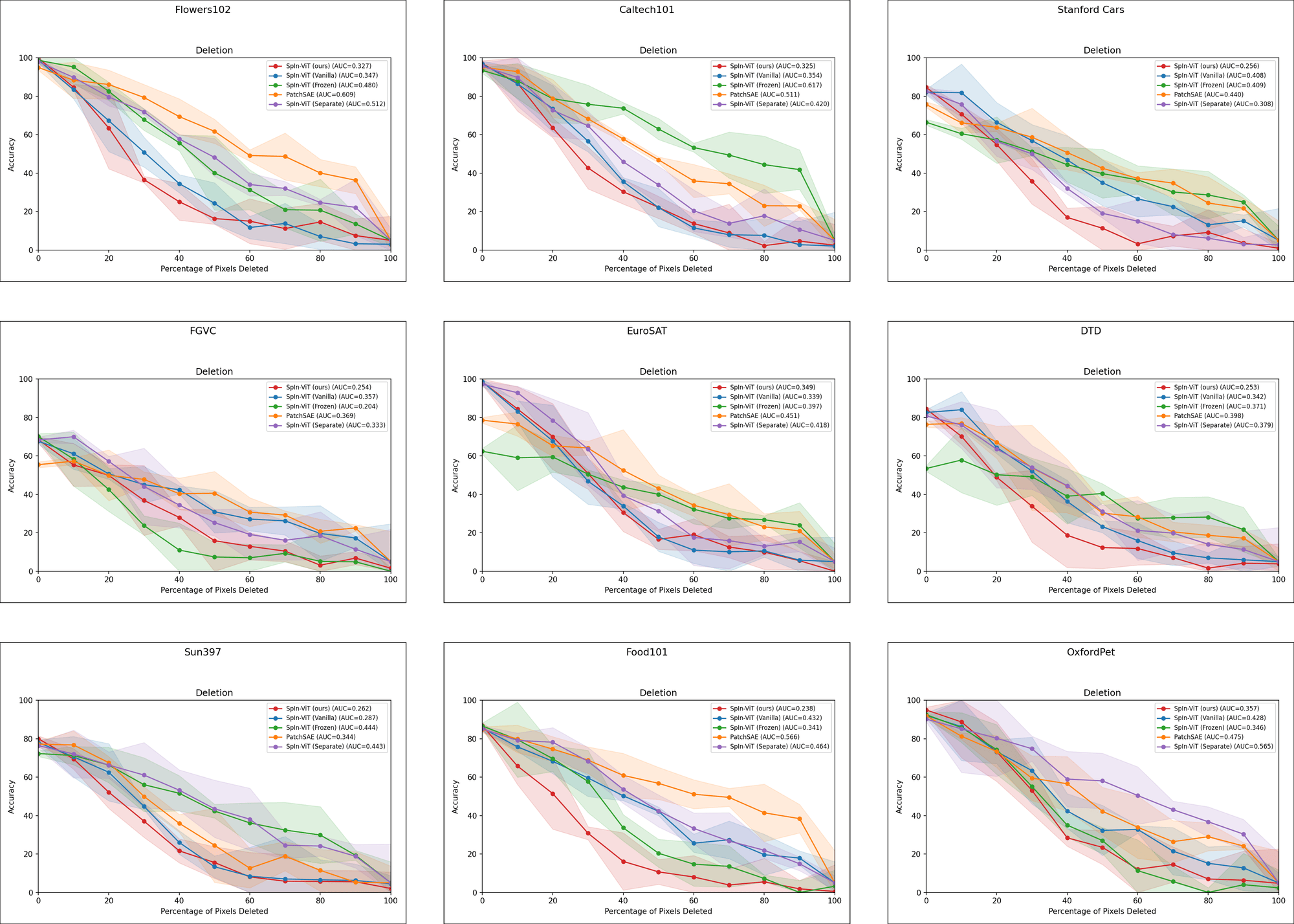}
   \caption{Deletion plots for all nine benchmark datasets. For each dataset, pixels are progressively removed from the original image in order of attribution ranking (most salient first, per each model's attribution map), and classification accuracy is plotted as a function of the percentage of pixels deleted. Steeper drops and lower AUC indicate more faithful attributions.}
   \label{fig:delete}
\end{figure*}

Figure~\ref{fig:delete} shows the complementary trend. SpIn-ViT (ours) exhibits the steepest early drop and lowest deletion AUC on most datasets, particularly Flowers102~\cite{nilsback2008automated}, Caltech101~\cite{4756141}, EuroSAT~\cite{helber2019eurosat}, and DTD~\cite{cimpoi2014describing}, confirming the regions we mark as important are also necessary for correct classification. SpIn-ViT (Frozen) and PatchSAE~\cite{lim2025patchsae} degrade more gradually, consistent with weaker insertion performance and more diffuse attributions. Separation between models is again smallest on FGVC~\cite{maji2013fine}, Sun397~\cite{xiao2010sun}, and Food101~\cite{bossard2014food}, where classification depends on a broader, less concentrated set of regions.

\subsubsection{Rule Extraction and Correspondence to Soft Patch Segmentation}
\label{sec:rules}
The symbolic rules extracted through FOLD-SE-M~\cite{wang2022foldse} work directly with the SAE's own feature space, rather than functioning as an explanation layer bolted on after training. Each predicate corresponds to a single sparse code dimension, so a rule such as \texttt{label(X, 'Truck') :- neuron\_63(X), not ab1(X)} is a logical condition over which SAE neurons are active, expressed in readable form rather than as a dense activation vector. Because the SAE is trained jointly with the classification objective, the selected neurons already drive the model's prediction, rather than being extracted post-hoc from an unrelated probe trained separately. This also means the rules inherit the sparsity of the representation, with a compact rule of two or three predicates sufficient to characterize a class, mirroring the sparsity constraint enforced during training.

\begin{figure*}[p]
   \includegraphics[width=1.0\textwidth]{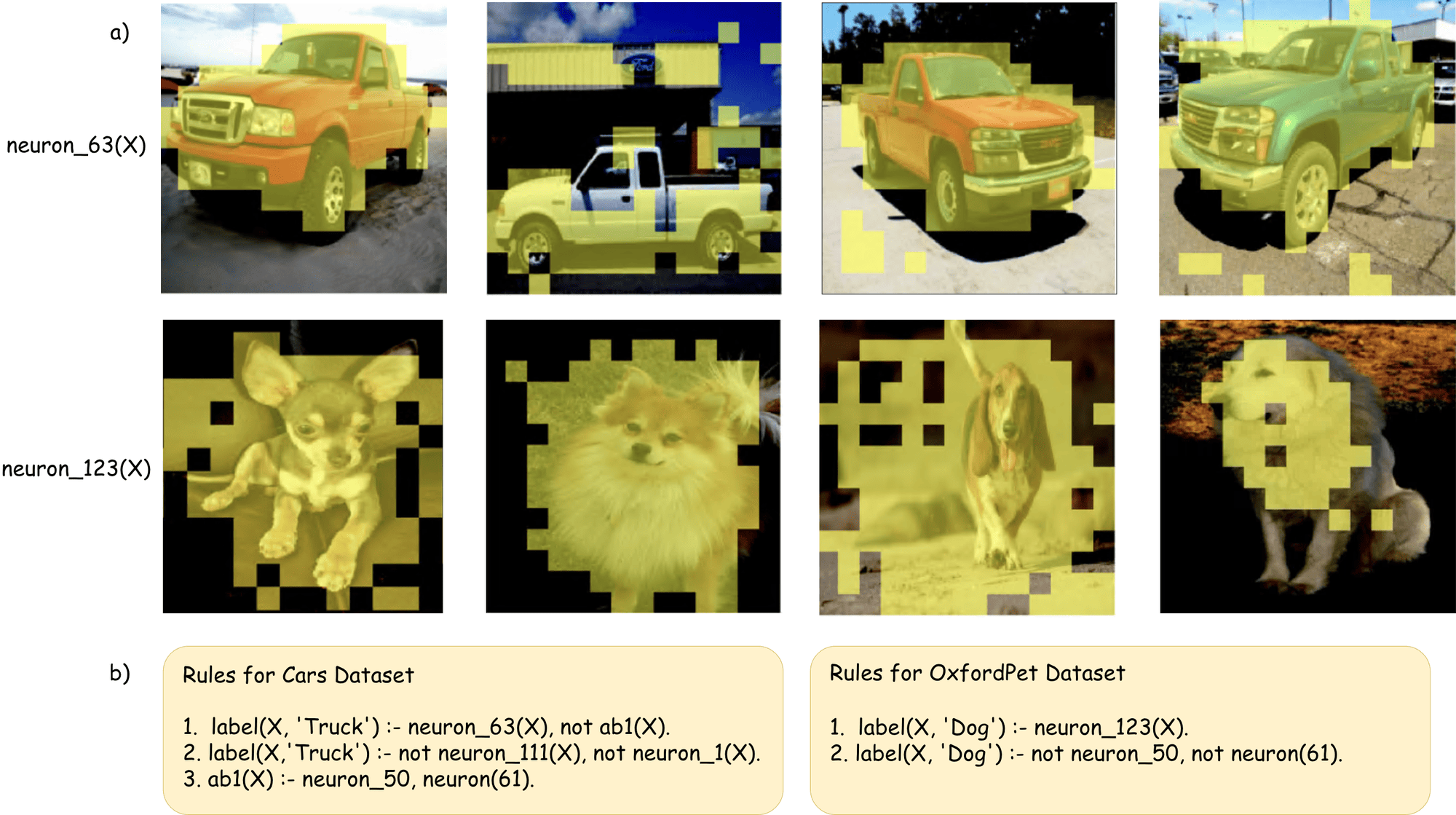}
   \caption{Examples of correspondence between sparse patch-level neuron activations and extracted symbolic rules. (a) Soft patch segmentations induced by individual neurons across example images. (b) FOLD-SE-M rules for the corresponding datasets, where each predicate maps to one of the visualized neurons.}
   \label{fig:rules}
\end{figure*}

Figure~\ref{fig:rules} illustrates this directly. Panel (a) shows the soft patch segmentation induced by neuron\_63 and neuron\_123, overlaying activation strength on each input image. Neuron\_63 activates over the body and cab of pickup trucks across Stanford Cars images regardless of color or orientation, while neuron\_123 activates over the head and body of the dog across OxfordPet images regardless of breed or pose. Panel (b) shows the corresponding FOLD-SE-M~\cite{wang2022foldse} rules: \texttt{label(X, 'Truck') :- neuron\_63(X), not ab1(X)} fires when the panel (a) segmentation activates over a truck-like region, with \texttt{ab1(X) :- neuron\_50(X), neuron\_61(X)} carving out exceptions where competing neurons also activate on the same image. For OxfordPet, \texttt{label(X, 'Dog') :- neuron\_123(X)} is satisfied whenever neuron\_123's segmentation covers the dog. In both cases, the rule and segmentation are two views of the same computation: the rule specifies which neurons must or must not fire, and panel (a) shows exactly where that firing occurs in the image.

\newpage \clearpage
\subsection{Example Labelled Rulesets}
\label{sec:rulesets}
{\scriptsize
\noindent\textbf{Flowers102} \{rose, sunflower, tulip, daisy, orchid\}:
\begin{verbatim}
label(X,'rose') :- neuron_63(X), not ab1(X).
label(X,'sunflower') :- neuron_142(X).
label(X,'tulip') :- not neuron_87(X), neuron_29(X).
label(X,'daisy') :- neuron_205(X), not neuron_142(X).
label(X,'orchid') :- neuron_11(X).
ab1(X) :- neuron_87(X), neuron_29(X).
\end{verbatim}

\noindent\textbf{Caltech101} \{airplane, chair, elephant, camera, piano\}:
\begin{verbatim}
label(X,'airplane') :- neuron_58(X), not ab2(X).
label(X,'chair') :- neuron_133(X).
label(X,'elephant') :- neuron_94(X), not neuron_133(X).
label(X,'camera') :- neuron_17(X).
label(X,'piano') :- not neuron_58(X), neuron_201(X).
ab2(X) :- neuron_9(X), neuron_112(X).
\end{verbatim}

\noindent\textbf{Stanford Cars} \{truck, sedan, convertible, SUV, coupe\}:
\begin{verbatim}
label(X,'truck') :- neuron_63(X), not ab1(X).
label(X,'truck') :- not neuron_111(X), not neuron_1(X).
label(X,'sedan') :- neuron_78(X).
label(X,'convertible') :- neuron_46(X), not neuron_78(X).
label(X,'SUV') :- neuron_63(X), neuron_190(X).
label(X,'coupe') :- not neuron_46(X), neuron_22(X).
ab1(X) :- neuron_50(X), neuron_61(X).
\end{verbatim}

\noindent\textbf{FGVC (Aircraft)} \{Boeing\_737, Airbus\_A320, Cessna\_172, Embraer\_E190\}:
\begin{verbatim}
label(X,'Boeing_737') :- neuron_84(X), not ab3(X).
label(X,'Airbus_A320') :- neuron_39(X).
label(X,'Cessna_172') :- not neuron_84(X), neuron_16(X).
label(X,'Embraer_E190') :- neuron_39(X), neuron_120(X).
ab3(X) :- neuron_16(X), neuron_120(X).
\end{verbatim}

\noindent\textbf{EuroSAT} \{Forest, Highway, Residential, River, AnnualCrop\}:
\begin{verbatim}
label(X,'Forest') :- neuron_27(X), not neuron_101(X).
label(X,'Highway') :- neuron_101(X).
label(X,'Residential') :- neuron_66(X), not ab4(X).
label(X,'River') :- not neuron_27(X), neuron_53(X).
label(X,'AnnualCrop') :- neuron_140(X).
ab4(X) :- neuron_53(X), neuron_140(X).
\end{verbatim}

\noindent\textbf{DTD (Textures)} \{striped, dotted, cracked, woven, grid\}:
\begin{verbatim}
label(X,'striped') :- neuron_72(X), not ab5(X).
label(X,'dotted') :- neuron_198(X).
label(X,'cracked') :- not neuron_72(X), neuron_35(X).
label(X,'woven') :- neuron_60(X), neuron_198(X).
label(X,'grid') :- not neuron_35(X), neuron_88(X).
ab5(X) :- neuron_60(X), neuron_88(X).
\end{verbatim}

\noindent\textbf{Sun397} \{bedroom, kitchen, office, forest\_road, beach\}:
\begin{verbatim}
label(X,'bedroom') :- neuron_14(X), not neuron_83(X).
label(X,'kitchen') :- neuron_41(X).
label(X,'office') :- neuron_122(X), not ab6(X).
label(X,'forest_road') :- not neuron_18(X), neuron_97(X).
label(X,'beach') :- neuron_206(X).
ab6(X) :- neuron_97(X), neuron_18(X).
\end{verbatim}

\noindent\textbf{Food101} \{pizza, sushi, steak, salad, burger\}:
\begin{verbatim}
label(X,'pizza') :- neuron_55(X), not ab7(X).
label(X,'sushi') :- neuron_129(X).
label(X,'steak') :- not neuron_55(X), neuron_74(X).
label(X,'salad') :- neuron_162(X), neuron_129(X).
label(X,'burger') :- not neuron_74(X), neuron_9(X).
ab7(X) :- neuron_162(X), neuron_9(X).
\end{verbatim}

\noindent\textbf{OxfordPet} \{Dog, Cat, Beagle, Persian, Labrador\}:
\begin{verbatim}
label(X,'Dog') :- neuron_123(X).
label(X,'Dog') :- not neuron_50(X), not neuron_61(X).
label(X,'Cat') :- neuron_88(X), not ab8(X).
label(X,'Beagle') :- neuron_123(X), neuron_31(X).
label(X,'Persian') :- not neuron_88(X), neuron_67(X).
label(X,'Labrador') :- neuron_123(X), not neuron_31(X).
ab8(X) :- neuron_67(X), neuron_31(X).
\end{verbatim}
}

\clearpage


\end{document}